\documentclass[journal]{IEEEtran}
\usepackage{graphicx}

\usepackage[english]{babel}
\usepackage{float}
\usepackage{makecell}
\usepackage{multirow}
\usepackage{cite}
\usepackage{amsmath,amssymb,amsfonts}
\usepackage{mathtools}
\usepackage{textcomp}

\usepackage{mathrsfs}

\usepackage{epstopdf}
\usepackage{comment}

\usepackage{enumerate}
\usepackage{booktabs}

\makeatletter
\newif\if@restonecol
\makeatother

\usepackage[linesnumbered,ruled,vlined]{algorithm2e}%[ruled,vlined]{
\usepackage{algpseudocode}
\begin{document}

\title{3D Scene Flow Estimation on Pseudo-LiDAR: Bridging the Gap on Estimating Point Motion}

\author{Chaokang Jiang, Guangming Wang, \IEEEmembership{Graduate Student Member, IEEE}, Yanzi Miao, and Hesheng Wang, \IEEEmembership{Senior Member, IEEE}
\thanks{This work was supported by the Fundamental Research Funds for the Central Universities 2020ZDPY0303. The first two authors contributed equally. Corresponding authors: Yanzi Miao; Hesheng Wang.}
\thanks{C. Jiang and Y. Miao is with Engineering Research Center of Intelligent Control for Underground Space, Ministry of Education, School of Information and Control Engineering, Advanced Robotics Research Center, China University of Mining and Technology, Xuzhou 221116, China.}
\thanks{G. Wang and H. Wang are with Department of Automation, Key Laboratory of System Control and Information Processing of Ministry of Education, Key Laboratory of Marine Intelligent Equipment and System of Ministry of Education, Shanghai Engineering Research Center of Intelligent Control and Management, Shanghai Jiao Tong University, Shanghai 200240, China.}}

% The paper headers
\markboth{Journal of \LaTeX\ Class Files,~Vol.~14, No.~8, August~2015}%
{Shell \MakeLowercase{\textit{et al.}}: Bare Demo of IEEEtran.cls for IEEE Journals}

\maketitle

\begin{abstract}
3D scene flow characterizes how the points at the current time flow to the next time in the 3D Euclidean space, which possesses the capacity to infer autonomously the non-rigid motion of all objects in the scene. The previous methods for estimating scene flow from images have limitations, which split the holistic nature of 3D scene flow by estimating optical flow and disparity separately. Learning 3D scene flow from point clouds also faces the difficulties of the gap between synthesized and real data and the sparsity of LiDAR point clouds. In this paper, the generated dense depth map is utilized to obtain explicit 3D coordinates, which achieves direct learning of 3D scene flow from 2D images. The stability of the predicted scene flow is improved by introducing the dense nature of 2D pixels into the 3D space. Outliers in the generated 3D point cloud are removed by statistical methods to weaken the impact of noisy points on the 3D scene flow estimation task. Disparity consistency loss is proposed to achieve more effective unsupervised learning of 3D scene flow. The proposed method of self-supervised learning of 3D scene flow on real-world images is compared with a variety of methods for learning on the synthesized dataset and learning on LiDAR point clouds. The comparisons of multiple scene flow metrics are shown to demonstrate the effectiveness and superiority of introducing pseudo-LiDAR point cloud to scene flow estimation.
\end{abstract}


\IEEEpeerreviewmaketitle

\section{Introduction}\label{Introduction}
%In the process of supervised or self-supervised training, a penalty signal is generated according to the estimation error and the network is optimized through back propagation of the penalty signal. This penalty signal may be too small due to the difference in distance information between the 2D space and the 3D space mentioned above. AAASelf-supervised learning of 3D scene flow from real-world images by our method outperforms that from synthesized datasets. In addition, the superiority of self-supervised learning of 3D scene flow from pseudo-LiDAR point clouds is demonstrated in the experiments compared to LiDAR point clouds.  \textcolor[RGB]{0,0,220}{} 3D scene flow can be projected onto an image to form an optical flow, which as a fundamental task in computer vision only characterizes 2D motion vectors.
\IEEEPARstart{T}{he} scene flow \cite{ref36} estimates the motion of 3D point in the scene, which is different from LiDAR odometry \cite{ref5} that estimates the consistent pose transformation of the entire scene. 3D scene flow is more flexible. The flexibility of scene flow makes it capable of assisting in many different tasks, such as object tracking \cite{ref1} and LiDAR odometry\cite{ref5}.

Generally, depth and optical flow together represent the scene flow in the scene flow estimation method based on 2D image \cite{ref11,ref10} or RGBD image\cite{teed2021raft}. Mono-SF \cite{ref11} infers the scene flow of visible points using constraints on motion invariance of multi-view geometric and depth distribution of a single view. To obtain more reliable scene flow, Hur et al. \cite{ref10} propagate temporal constraints to continuous multi-frame images. RAFT-3D \cite{teed2021raft} estimates the scene flow by soft grouping pixels into rigid objects. The nature of the scene flow being the 3D motion vectors is split by the methods for estimating scene flow by optical flow and pixel depth in image space. Because of the lack of explicit 3D geometry information in the image space, these methods often cause large pixel matching errors during scene flow inference. For example, two unrelated points that are far apart in 3D space may be very close to each other in the image plane. Dewan et al. \cite{ref21} predict 3D scene flow from adjacent frames of LiDAR data in a without-learning method. This method requires various strong assumptions, such as that the local structure will not be deformed by the motion in the 3D scene. Some recent works \cite{ref4,ref16,ref19,ref27} learn 3D scene flow from point cloud pairs based on deep neural network. However, these methods of 3D scene ﬂow estimation are invariably self-supervised/supervised on the synthesized dataset FlyingThings3D \cite{ref24} and evaluated the generalization of the model on the real-world dataset KITTI Scene Flow \cite{sfkitti}. The models trained on the synthesized dataset will cause accuracy degradation in the real world \cite{ref16}. 
%In addition, the synthesized dataset will ignore some random behaviors to destroy the generalizability of the network. Learning 3D scene flow on real-world datasets is considered more suitable for practical applications. Mittal et al. \cite{ref27} present two self-supervised losses, nearest neighbor loss, and cycle consistency loss. PointPWC-Net \cite{ref16} introduces three self-supervised losses, Chamfer distance, smoothness constraint, and Laplacian regulation. Following the PointPWC-Net setup, we train the scene flow estimator on LiDAR point cloud data, and the evaluation results demonstrate numerous cases of inference failure. The 3D scene flow estimation process is to find the soft correspondence between adjacent frame points by matching the spatial characteristics of the points.

\begin{figure}[t]
	\centering
	\includegraphics[scale=0.26]{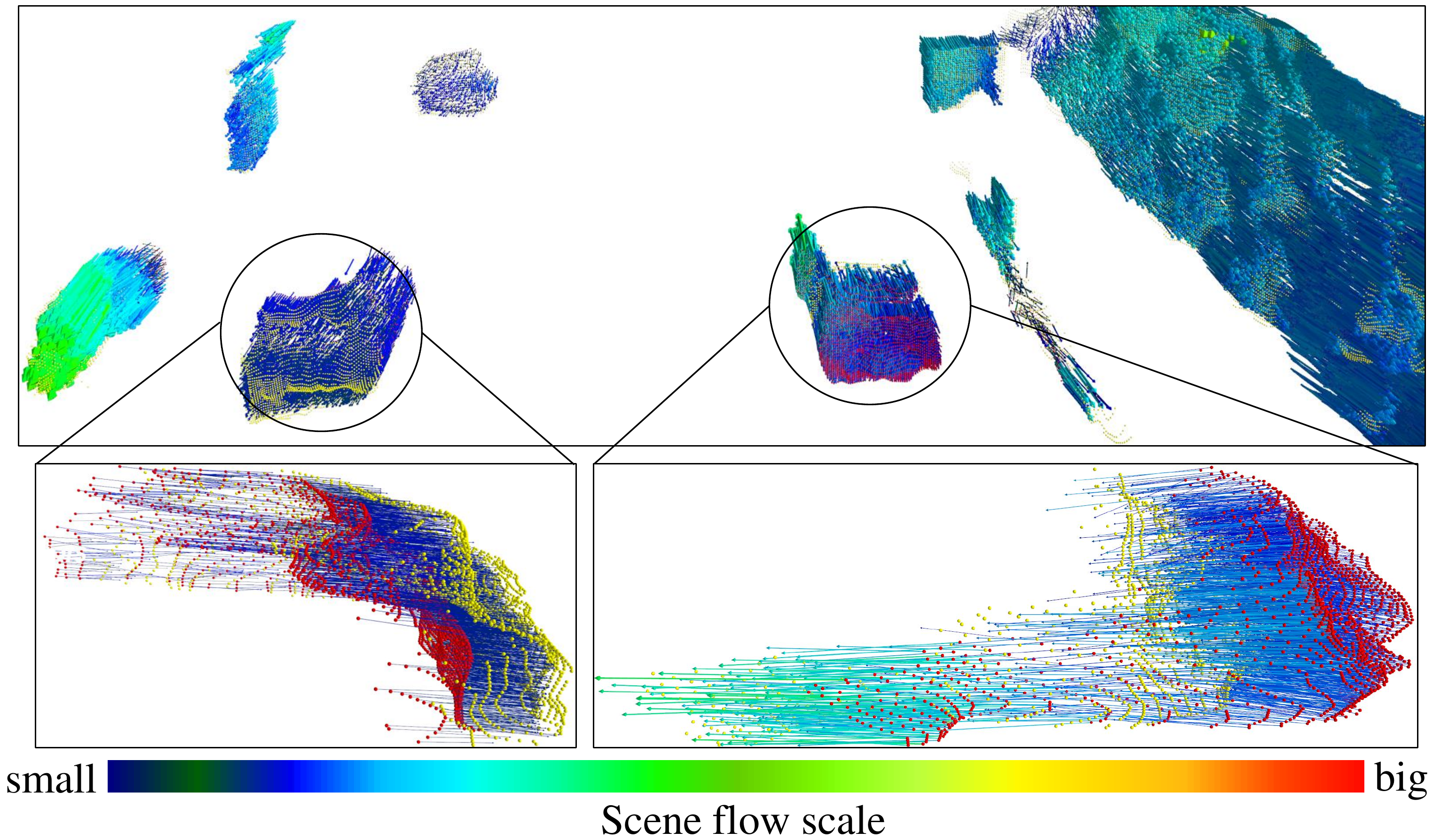}
	%\vspace{-5mm}
	\caption{\textbf{3D scene flow estimation from 2D images.} The first frame and the second frame are red points and yellow points, respectively. The 3D scene flow vector is represented as arrows of different colors. The low-to-high color map on the bottom shows the estimated scene flow vector from small to large.}
	\label{fig:Connect}
\end{figure}

3D scene flow annotations are very scarce in real-world datasets. Some works \cite{ref27, ref16} proposes some excellent self-supervised losses, but these are difficult to achieve success on LiDAR signals. LiDAR signals are recognized to have two weaknesses, sparsity and point cloud distortion. Firstly, the existing self-supervised losses \cite{ref16,ref27} imply a strong assumption of point-by-point correspondence. But point clouds from different moments are inherently discrete. The point-by-point correspondence and the discrete nature of the point cloud are contradictory. The sparsity of the LiDAR signal further exacerbates this fact. Secondly, the data collection process for mechanical LiDAR is accompanied by motion, which results in point cloud points in the same frame not being collected at the same moment, i.e., point cloud distortion. However, 3D scene flow estimation is the process of local 3D matching, which requires accurate temporal information. Unlike the LiDAR signal, the pseudo-LiDAR signal comes from the back-projection of the dense depth map into a 3D point cloud. Almost no distortion is caused by pseudo-LiDAR due to the instantaneous capture of the image. A novel structure is designed in this paper to enable self-supervised learning of scene flow to benefit from pseudo-LiDAR. Although the spatial position of the points in the point cloud generated from the depth map is not always accurate, our method is still able to find the correspondence between the points in adjacent frames well. As shown in Fig. \ref{fig:Connect}, the point clouds of adjacent frames have similar representations of same objects in the scene because the performance of the depth estimation network is unchanged. Thus the spatial features of a point at frame $t$ can easily find similar spatial features at frame $t+1$ and this match is usually accurate.
%The color camera used in the KITTI dataset costs no more than \$800, while the Velodyne HDL-64E LiDAR sensor costs around \$75,000 \cite{ref30}. Recently, Ouster published an inexpensive OS1-128 LiDAR that still costs \$18,000 \cite{durlar}.}The cost of hardware for autonomous driving or robotics will be greatly reduced as a result. \textcolor[RGB]{0,0,220}{

On the other hand, the camera is cheaper than LiDAR. Although the cost of LiDAR is shrinking year by year, there is still an undeniable cost difference between LiDAR and camera. Recently, Ouster released an inexpensive OS1-128 LiDAR, but the price still costs \$18,000 \cite{durlar}. The stereo camera system may provide a low-cost backup system for the LiDAR-based scheme of scene flow estimation.

In summary, our key contributions are as follows:

\begin{itemize}
\item A self-supervised method for learning 3D scene flow from stereo images is proposed. Pseudo-LiDAR based on stereo images is introduced into 3D scene flow estimation. The method in this paper bridges the gap between 2D data and the task of estimating 3D point motion. 
\item The sparse and distortion characteristics of the LiDAR point cloud bring errors for the calculation of existing self-supervised loss of the 3D scene flow. The introduction of the pseudo-LiDAR point cloud in this paper improves the effectiveness of these self-supervised losses because of the dense and non-distortion characteristics of the pseudo-LiDAR point cloud, which is demonstrated by experiments in this paper.
\item 3D points with large errors caused by depth estimation are filtered out as much as possible to reduce the impact of noise points on the scene flow estimation. A novel disparity consistency loss is proposed by exploiting the coupling relationship between 3D scene flow and stereo depth estimation.
\end{itemize}

% The experiments show that the accuracy of our method for estimating 3D scene flow is better than that of LiDAR point cloud-based and synthesized point cloud-based methods.
%There are five sections to describe the studies in this paper. Section \ref{Introduction} and Section \ref{RelatedWork} are introduction and related work, respectively. Section \ref{Overview} describes the details of our method. Experimental details and visualization are presented in Section \ref{Experiments}. Conclusions are summarized in Section \ref{Conclusion}.
%They introduce ``Change of Representation (CoR)'' module which allows the object detection loss to be backpropagated to the depth estimation network and allows end-to-end training of both networks. Weng et al. \cite{ref32} train a 3D object detection network using pseudo-LiDAR from monocular depth estimation and reduce the noise of pseudo-LiDAR using the constraint of the 2D-3D bounding box consistency and instance masks instead of bounding boxes.  Lv et al. \cite{ref13} estimate 3D motion fields based on RGBD information by inferring rigidity, poses, and 2D optical flow in an attempt of eliminating camera motion ambiguities. 

\section{Related Work}\label{RelatedWork}

\subsection{Pseudo-LiDAR}
In recent year, many works \cite{ref29,ref30,ref31} build pseudo-LiDAR based 3D object detection pipeline. Pseudo-LiDAR has shown significant advantages in the field of object detection. Wang et al. \cite{ref29} introduce pseudo-LiDAR into an existing LiDAR object detection model and demonstrate that the main reason for the performance gap between stereo and LiDAR is the representation of the data rather than the quality of the depth estimate. Pseudo-LiDAR++ \cite{ref30} optimizes the structure and loss function of the depth estimation network based on Wang et al. \cite{ref29} to enable the pseudo-LiDAR framework to accurately estimate distant objects. Qian et al. \cite{ref31} build on Pseudo-LiDAR++ \cite{ref30} to address the problem that the depth estimation network and the object detection network must be trained separately. The previous pseudo-LiDAR framework focuses more on scene perception with the single frame, while this paper focuses on the motion relationship between two frames. 
% In the autonomous driving scene, the 3D scene flow of dynamic objects with different speeds and directions can be accurately estimated by our model from 2D images only.

\subsection{Scene Flow Estimation}

Some works study the estimation of dense scene flow from images of consecutive frames. Mono-SF \cite{ref11} proposes ProbDepthNet to estimate the pixel depth distribution of the single image. Geometric information from multiple views and depth distribution information from a single view are used to jointly estimate the scene flow. Hur et al. \cite{ref10} introduce a multi-frame temporal constraint to the scene flow estimation network. Chen et al. \cite{ref36} develop a coarse-grained software framework for scene-flow methods and realize real-time cross-platform embedded scene flow algorithms. In addition, Rishav et al. \cite{ref3} fuse LiDAR and images to estimate dense scene flow. But they still perform feature fusion in image space. These methods rely on 2D representations and cannot learn geometric motion from explicit 3D coordinates. The pseudo-LiDAR is the bridge between the 2D signal and the 3D signal, which provides the basis for directly learning the 3D scene flow from the 2D data.
\begin{figure*}[t]
    %\vspace{-2mm}
	\includegraphics[scale=0.56]{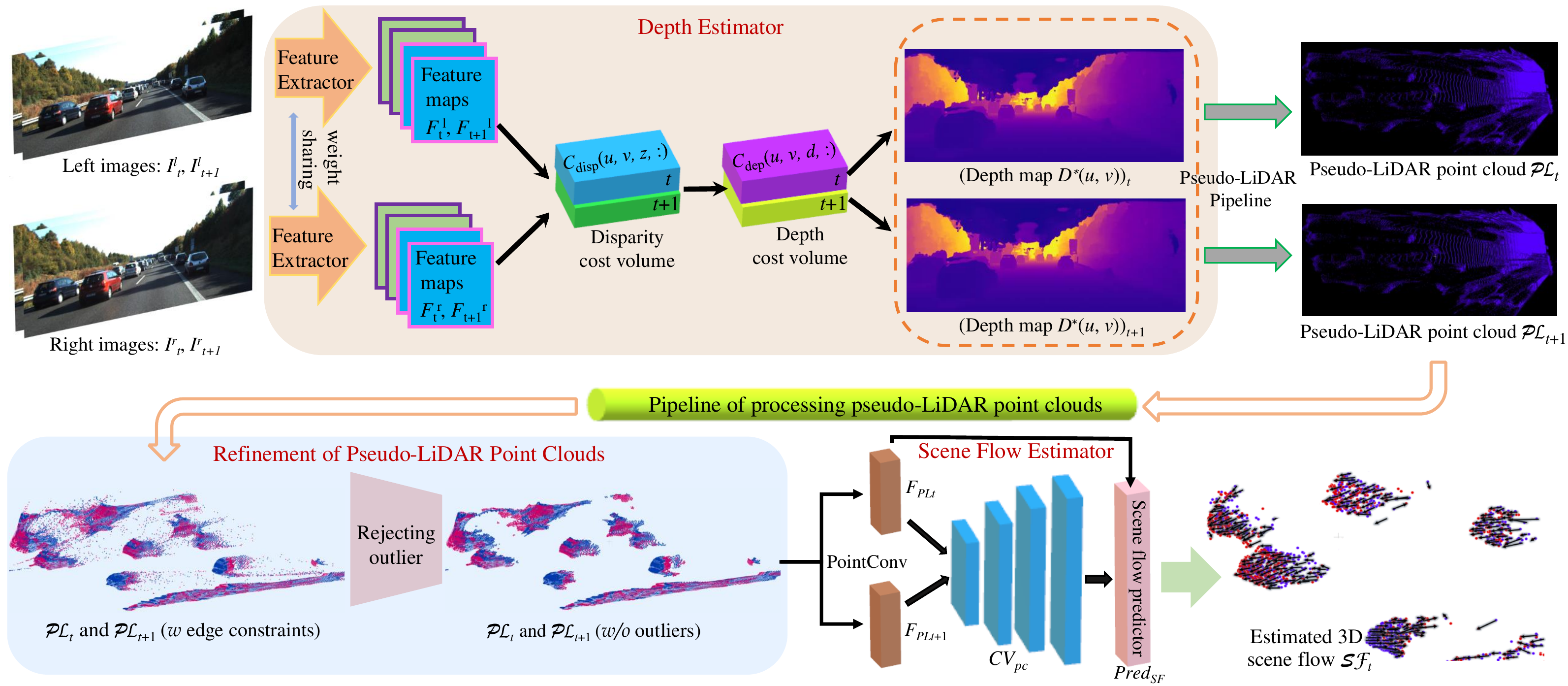}
	\caption{\textbf{Overview of the framework for estimating 3D scene flow from stereo images.} Depth estimator workflow is described in section \ref{DepthEst}. Pseudo-LiDAR pipeline is described in detail in section \ref{Overview}. Refinement of Pseudo-LiDAR Point Clouds is described in section \ref{Refinementpc}, where the red and blue points represent the point clouds at frame t and frame t+1, respectively. ``$w$ edge constraints'' denotes the operation of filtering edge error points of the pseudo-LiDAR point cloud, which is explained specifically in section \ref{Edges}. ``$w/o$ outliers'' denotes the operation of filtering outliers, which is explained specifically in section \ref{Remove:Outliers}. Scene flow estimator is described in section \ref{SFestimation}. Pseudocode for the overall framework is also provided in this paper in Algorithm \ref{algorithm1}.}
	\label{fig:Framework}
\end{figure*}

The original geometric information is preserved in the 3D point cloud, which is the preferred representation for many scene understanding applications in self-driving and robotics. Some researchers \cite{ref21} estimate 3D scene flow from LiDAR point clouds by using the classical method. Dewan et al. \cite{ref21} introduce local geometric constancy during the motion and introduce a triangular grid to determine the relationship of the points. Benefiting from the point cloud deep learning, some recent works \cite{ref4,ref16,ref19,ref27} propose to learn 3D scene flow from raw point clouds. FlowNet3D \cite{ref4} firstly proposes the flow embedding layer which finds point correspondence and implicitly represents the 3D scene flow. FLOT \cite{ref19} studies lightweight structures for optimizing scene flow estimation using optimal transport modules. PointPWC-Net \cite{ref16} proposes novel learnable cost volume layers to learn 3D scene flow in a coarse-to-fine approach, and introduces three self-supervised losses to learn the 3D scene flow without accessing the ground truth. Mittal et al. \cite{ref27} propose two new self-supervised loss.
%The total number of model parameters for FLOT is 0.11 million. In contrast, PointPWC-Net contains 7.7 million parameters. Self-Point-Flow \cite{ref20} utilizes multiple descriptors to design transport cost in optimal transport theory, constraining high quality one-to-one matching.
\section{Self-Supervised Learning of the 3D Scene Flow from Pseudo-LiDAR}

\subsection{Overview}\label{Overview}
\begin{algorithm}
\label{algorithm1}
  \caption{General Scheme of Learning Scene Flow}
  \KwIn{Stereo images \{($I_{t}^l, I_{t}^r$), ($I_{t+1}^l, I_{t+1}^r$) $|$ $t = 1, 2, ..., M$\}, camera intrinsic $C$.}
  \KwOut{3D scene flow \{$\mathcal{SF}_1, \mathcal{SF}_2, ..., \mathcal{SF}_M$\}}
  \While{not convergent}
  {
   Feature maps \{($F_{t}^l$, $F_{t}^r$), ($F_{t+1}^l$, $F_{t+1}^r$)\} are extracted from \{($I_{t}^l$, $I_{t}^r$), ($I_{t+1}^l$, $I_{t+1}^r$)\} by CNN. \\
   Construct 4D depth cost volume $C^{t}_{dep}$, $C^{t+1}_{dep}$ based on \{($F_{t}^l$, $F_{t}^r$), ($F_{t+1}^l$, $F_{t+1}^r$)\}. \\
   Regress depth maps \{$D_t$, $D_{t+1}$\} and compute \{$\mathcal{PL}_{t}$, $\mathcal{PL}_{t+1}$\} with Eq. (\ref{equ:projected}).\\
   Filter the boundary points in \{$\mathcal{PL}_{t}$, $\mathcal{PL}_{t+1}$\} and filter the outliers with Eq. (\ref{eq:Out}).\\
   \For{$l_i$ in range(0, $L$)}
      {
      Encode \{$\mathcal{PL}_{t}$, $\mathcal{PL}_{t+1}$\} by PointConv to output point cloud features \{$F_{PL1}$, $F_{PL2}$\}.\\
      Construct $CV_{pc}$ based on \{$F_{PL_{t}}$, $F_{PL_{t+1}}$\}.\\
      Estimation of $\mathcal{SF}_t=\{sf_i|i=1,2,\cdots,n_1\}$ of $\mathcal{PL}_{t}$ using $Pred_{SF}$.\\
      }
    Update the scene flow estimator weight parameters $\theta_2$ by minimizing the objective in Eq. (\ref{eq:Total_loss}).
  }
\end{algorithm}
The main purpose of this paper is to recover 3D scene flow from stereo images. The stereo images are represented by $I^l$ and $I^r$ respectively. As Fig. \ref{fig:Framework}, given a pair of stereo images, which contains reference frames $\{I_{t}^l, I_{t}^r\}$ and target frames $\{I_{t+1}^l, I_{t+1}^r\}$. Each image is represented by a matrix of dimension $H\times W\times 3$. Depth map $D_t$ at time $t$ is predicted by feeding the stereo image $\{I_{t}^l, I_{t}^r\}$ into the depth estimation network $D_{net}$. Each pixel value of $D$ represents the distance $d$ between a certain point in the scene and the left camera. Pseudo-LiDAR point cloud comes from back-projecting the generated depth map to a 3D point cloud, as follow:
\begin{equation}
    \left\{
   \begin{array}{c}
   x_w = d \times (u - c_x) \times \frac{1}{f_x}\\  
   y_w = d \times (v - c_y) \times \frac{1}{f_y}\\ 
   z_w = d ,
   \end{array}
\right.
\label{equ:projected}
\end{equation}
where $f_x$ and $f_y$ are the horizontal and vertical focal lengths of the camera, and $c_x$ and $c_y$ are the coordinate center of the image, respectively. The 3D point coordinate $(x_w, y_w, z_w)$ in the pseudo-LiDAR point cloud $\mathcal{PL}$ is calculated by pixel coordinates $(u, v)$, $d$ and camera intrinsics. $\mathcal{PL}_t = \{c_{1,i}\in{\mathbb{R}^3}\}_{i=1}^{N_1}$ with $N_1$ points and $\mathcal{PL}_{t+1} = \{c_{2,j}\in{\mathbb{R}^3}\}_{j=1}^{N_2}$ with $N_2$ points are generated from the depth maps $\mathcal{D}_t$ and $\mathcal{D}_{t+1}$, where $c_{1,j}$ and $c_{2,j}$ are the 3D coordinates of the points. $\mathcal{PL}_t$ and $\mathcal{PL}_{t+1}$ are randomly sampled to $N$ points, respectively. The sampled pseudo-LiDAR point clouds are passed into the scene flow estimator $\mathcal{F}_{sf}$ to extract the scene flow vector $\mathcal{SF}_t=\{sf_i|i=1,2,\cdots,N\}$ for each 3D point in frame $t$.
\begin{equation}
SF_t = \mathcal{F}_{sf} \{\mathcal{P}[D_{net}(I_t; \theta_1)], \mathcal{P}[D_{net}(I_{t+1}; \theta_1)]; \theta_2\},
\end{equation}
where $\theta_1$ and $\theta_2$ are the parameters of the network. $\mathcal{P}$ represents the back-projection by Eq. (\ref{equ:projected}). 

It is difficult  to obtain the ground truth scene flow of pseudo-LiDAR point clouds. Mining a priori knowledge from the scene itself to self-supervised learning of 3D scene flow is essential. Ideally, 
$\mathcal{PL}_{t+1}$ and estimated point cloud $\mathcal{PL}_{t+1}^{\omega}$ have the same structure.
With this priori knowledge, point cloud $\mathcal{PL}_{t}$ is warped to point cloud $\mathcal{PL}_{t+1}^{\omega}$ through the predicted scene flow $\mathcal{SF}_t$,
\begin{equation}
\mathcal{PL}_{t+1}^{\omega} = \mathcal{PL}_{t} + \mathcal{SF}_t.
\end{equation}
Based on the consistency of $\mathcal{PL}_{t+1}$ and $\mathcal{PL}_{t+1}^{\omega}$, the loss functions Eq. (\ref{eq:chamfer}) and Eq. (\ref{eq:laplace}) are utilized to implement self-supervised learning. We provide the pseudocode in Algorithm \ref{algorithm1}  for our method, where $L$, $CV_{pc}$, and $Pred_{SF}$ are described in detail in Section \ref{SFestimation}.

\subsection{Depth Estimation}\label{DepthEst}
%\subsubsection{depth estimation}
%The predicted disparity $\widetilde{d}$ is obtained by weighting the probability of each disparity $\widetilde{d}$, where $\widetilde{d}$ is calculated from the probability cost volume $C_{disp}(u,v,d,:)$ through a softmax operation.Each pixel is processed equally by the 3D convolution in $C_{disp}(u,v,d,:)$.which allows a relatively accurate response for the motion, as well as the higher accuracy local pose estimation and the complete decoupling of the state estimation.The pixels of the estimated depth map are back-projected into 3D coordinates. The principle of the back-projection is described in Formula \ref{equ:projected}. Pseudo-LiDAR point cloud $\mathscr{PL}$ is generated with a small difference from the ground truth of the point cloud. We use some advantages of $\mathscr{PL}$ to compensate the disadvantages of LiDAR data in 3D scene flow estimation as much as possible.
The disparity $z$ is the horizontal offset of the corresponding pixel in the stereo image, which represents the difference caused by viewing the same object from a stereo camera. $I_l(u,v)$ and $I_r(u,v+z)$ represent the observation of the same 3D point in space. Two cameras are connected by a line called the baseline. The distance $d$ between the object and the observation point can be calculated by knowing the disparity $z$, the baseline length $b$, and the horizontal focal length $f$.
\begin{equation}
d = \frac{b \times f}{z}.
\label{eq:ZD}
\end{equation}
Disparity estimation networks such as PSMNet \cite{ref38} extract deep feature maps $\{F_{t}^l, F_{t}^r\}$ and $\{F_{t+1}^l, F_{t+1}^r\}$ from $\{I_{t}^l, I_{t}^r\}$ and $\{I_{t+1}^l, I_{t+1}^r\}$, respectively. As shown in Fig. \ref{fig:Framework}, the features of $F_{t}^l(u,v)$ and $F_{t}^r(u,v+z)$ are concatenated to construct 4D tensor $C_{disp}(u,v,z,:)$, namely the disparity cost volume. Then 3D tensor $H_{disp}(u,v,z)$ is calculated by feeding $C_{disp}(u,v,z,:)$ into the 3D convolutional neural network (CNN). The predicted pixel disparity $D^*(u,v)$ is calculated by softmax weighting $\sum\limits_{z}softmax(H_{disp}(u,v,z)) \times z$ \cite{ref38}. Based on the fact that disparity and depth are inversely proportional to each other, the convolution operation in the disparity cost volume has disadvantages. The same convolution kernel is applied to a few pixels with small disparity (i.e., large depth) resulting in an easy skipping and ignoring many 3D points. It is more reasonable to run the convolution kernel on the depth grid that produces the same effect on neighbor depths, rather than overemphasizing objects with large disparity (i.e., small depth) on the disparity cost volume. Based on this insight, the disparity cost volume $C_{disp}(u,v,z,:)$ is reconstructed as depth cost volume $C_{dep}(u,v,d,:)$ \cite{ref30}. Finally, the depth of the pixel is calculated through a similar weighting operation as mentioned above.

The sparse LiDAR points are projected onto the 2D image as the ground truth depth map $D_{gt}$. The depth loss is constructed by minimizing the depth error: 
\begin{equation}
\mathcal{L}_1 = \sum\limits_{(u,v) \in D_{gt}} \tau(D_{gt}(u,v) - D^*(u,v)).
\label{eq:Z_Z}
\end{equation}
$D^*$ represents the predicted depth map. $\tau$ represents smooth L1 loss.

%The previous methods of separately estimating depth and optical flow from 2D images to characterize scene flow have data gap. We use pseudo-LiDAR to directly estimate the 3D scene flow to bridge this gap. The utilization of pseudo-LiDAR also avoids the drawbacks of 3D scene flow estimation on LiDAR data. However, the s   Although outliers and sky points in the pseudo-LiDAR point cloud are filtered out, the point cloud still retains more than 300,000 3D points. However, a LiDAR point cloud with 360-degree view contains only about 110,000 3D points per frame.
\subsection{Refinement of Pseudo-LiDAR Point Clouds} \label{Refinementpc}
Scene flow estimator cannot directly benefit from the generated original pseudo-LiDAR point clouds due to its containing many points with estimation error.
Reducing the impact of these noise points on the 3D scene flow estimation is the problem to be solved.
\subsubsection{LiDAR and Pseudo-LiDAR for 3D Scene Flow Estimation} \label{l_pl}

LiDAR point clouds in KITTI raw dataset \cite{ref40} are captured using low-frequency (10 $Hz$) mechanical LiDAR scans. Each frame of the point cloud is collected via the rotating optical component of the LiDAR at low frequencies. This process is accompanied by the motion of the LiDAR itself and the motion of other dynamic objects. Therefore, the raw LiDAR point cloud contains a lot of distortion points. The point cloud distortion generated by its self-motion can be largely eliminated by collaborating with other sensors (e.g. inertial measurement unit). However, point cloud distortion that is caused by other dynamic objects is difficult to be eliminated, which is one of the challenges of learning 3D scene flow from LiDAR point clouds. In comparison, the sensor captures images with almost no motion distortion, which is an important advantage for recovering 3D scene flow from pseudo-LiDAR signals.

\begin{figure}[t]
	\centering
	\includegraphics[scale=0.40]{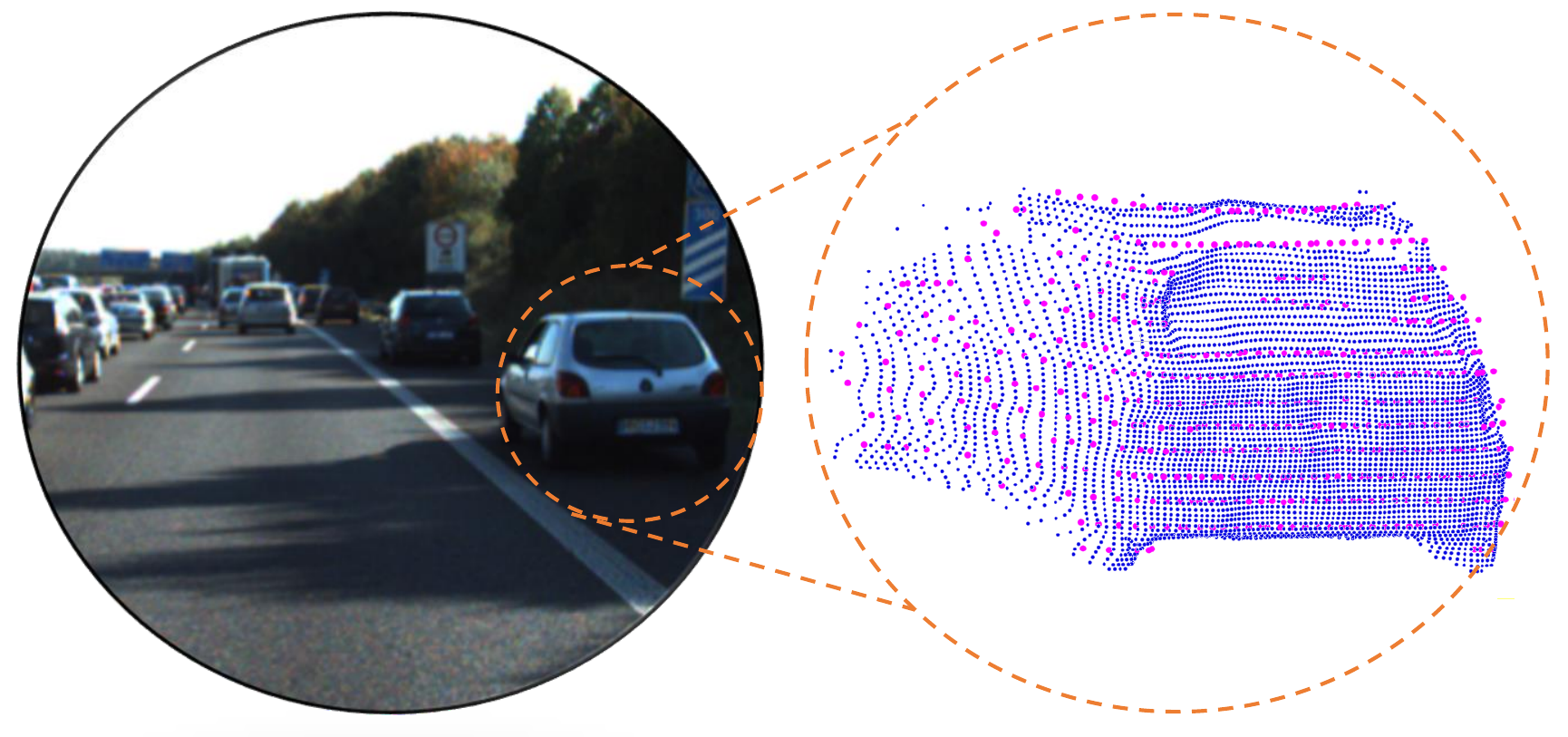}
	%\vspace{-1mm}
	\caption{\textbf{LiDAR point cloud and pseudo-LiDAR point cloud of the same object.}
	LiDAR points are zoomed in and are represented as red. Pseudo-LiDAR points are represented as blue. The pseudo-LiDAR point cloud is significantly denser than the LiDAR point cloud, and they are well aligned}
	\label{fig:pseudolidar}
\end{figure}

As shown in Fig. \ref{fig:pseudolidar}, the point cloud from LiDAR (64-beam) in KITTI dataset \cite{ref40} is sparsely distributed on 64 horizontal beams. In contrast,  pseudo-LiDAR point clouds come from dense pixel and depth values, which are inherently dense. The image size in KITTI is 1241 $\times$ 376. This means that the image contains 466,616 pixel points.  The self-supervised learning of 3D scene flow mentioned in subsection \ref{Overview} assumes that the warped point cloud $\mathcal{PL}_{t+1}^{\omega}$ and the point cloud $\mathcal{PL}_{t+1}$ correspond point by point. However, the disadvantage of this assumption is magnified by the sparse and discrete nature of the LiDAR point cloud. For example, Chamfer loss \cite{ref16} forces the point cloud of these two frames to correspond point by point, which makes self-supervised loss over-punish the network so that it does not converge to the global optimum. 
% Laplacian regularization loss \cite{ref16} also contains a point-by-point correspondence assumption similar to Chamfer loss. when both $\mathcal{PL}_{t+1}^{\omega}$ and $\mathcal{PL}_{t+1}$ characterize the same car surface with sparse points, To reduce the impact of noise points in the original pseudo-LiDAR point cloud for 3D scene flow estimation, these error points are found and eliminated. 
%{\color[RGB]{255, 3, 255} red}
\subsubsection{Constraints on Pseudo-LiDAR Point Cloud Edges} \label{Edges}

A large number of incorrectly estimated points are distributed at the scene boundaries. For example, the generated pseudo-LiDAR point cloud has long tails on the far left and far right. Weakly textured areas such as white sky also result in a lot of depth estimation errors. Appropriate boundaries are specified for pseudo-LiDAR point clouds to remove as many edge error points as possible and not to lose important structural information.

\subsubsection{Remove Outliers from the Pseudo-LiDAR Point Cloud} \label{Remove:Outliers}
It is also very important to remove the noise points inside the pseudo-LiDAR point cloud. As shown in Fig. \ref{fig:Framework}, a straightforward and effective method is to find outliers and eliminate them. In the lower-left corner of Fig. \ref{fig:Framework}, a long tail is formed on the edge of the car, and these estimated points have deviated from the car itself. Statistical analysis is performed on the neighborhood of each point. The average distance $\overline{v}_m$ from it to the $m$ neighboring points is calculated. The obtained result is assumed to be a Gaussian distribution, which shape is determined by the mean and standard deviation. This statistical method is useful to find discordant points in the whole point cloud. A point in the point cloud is considered as an outlier when its distance from its nearest point is larger than a distance threshold $d_{max}$:
\begin{equation}
d_{max} = \overline{v}_m + \alpha \times \sqrt{\frac{1}{m-1} \sum_{k=1}^m(v_k-\overline{v}_m)^2}.
\label{eq:Out}
\end{equation}
$d_{max}$ is determined by the scaling factor $\alpha$ and the number $m$ of nearest neighbors.

\subsection{3D Scene Flow Estimator}\label{SFestimation}

%The previous image-based scene flow estimation extracts features from the images by convolutional neural networks and finds the correspondence between the adjacent frame points according to these features. 
The image convolution process is the continuous multiplication and summation of the convolution kernel in the image space. This operation is flawed for matching points in real-world space. Points that are far apart in 3D space may be close together on a depth map, such as the edge of a building or the edge of a car. This problem cannot be paid attention to by convolution on the image, which leads to incorrect feature representation or feature matching. Convolution on 3D point clouds better avoids that flaw.

The generated pseudo-LiDAR point clouds $\mathcal{PL}_t$ and $\mathcal{PL}_{t+1}$ are encoded and downsampled by PointConv \cite{ref16} to obtain the point cloud features $F_{PL_t}=\{F^t_i|i=1,2,\cdots,n_1\}$ and $F_{PL_{t+1}}=\{F^{t+1}_i|i=1,2,\cdots,n_2\}$.
Then the matching cost between point $pl_{t,i}$ and point $pl_{t+1,j}$ is calculated by concatenating the features $F^{t}_i$ of $pl_{t,i}$, the features $F^{t+1}_i$ of $pl_{t+1,j}$, and the direction vector $pl_{t,i}-pl_{t+1,j}$ \cite{ref16}, where $pl_{t,j} \in \mathcal{PL}_{t}$ and $pl_{t+1,j} \in \mathcal{PL}_{t+1}$. The nonlinear relationship between $pl_{t,i}$ and $pl_{t+1,j}$ is learned by using multilayer perceptron. According to the obtained matching costs, the point cloud cost volume $CV_{pc}$ used to estimate the movement between points is aggregated. The scene flow estimator constructs a coarse-to-fine network for scene flow estimation. The coarse scene flow is estimated by feeding point cloud features and $CV_{pc}$ into the scene flow predictor $Pred_{SF}$. The input of $Pred_{SF}$ is the point cloud features of the first frame of the current level, $CV_{pc}$, scene flow $\mathcal{SF}_{l+1}^{up}$ from the last level of upsampling, and point cloud features from the last level of upsampling. The output is the scene flow and point cloud features of the current level. The local features of the four variables of $Pred_{SF}$ input are merged by using the PointConv \cite{ref16} layer and the new $N_l \times C$ dimensional features are output. The new $N_l \times C$ dimensional features are used as input to the  multilayer perceptron to predict the current level of scene flow. The final output is the 3D scene flow $\mathcal{SF}_t=\{sf_i|i=1,2,\cdots,n_1\}$ of each point in $\mathcal{PL}_t$.
%where the $Pred_{SF}$ consists of PointConv \cite{ref16} and MLP. Then the finer flow is predicted by using point cloud features, $CV_{pc}$, and the coarse flow from the upper level. The final output is the 3D scene flow $\mathcal{SF}_t=\{sf_i|i=1,2,\cdots,n_1\}$ of each point in $\mathcal{PL}_t$. disparity consistency

The self-supervised loss is used at each level to optimize the prediction of the scene flow. The proposed network has four unsupervised losses Chamfer loss $\mathcal{L}_C$, smoothness constraint $\mathcal{L}_{SC}$, Laplacian regularization $\mathcal{L}_{LR}$, and disparity consistency $\mathcal{L}_{DC}$. 
$\mathcal{PL}_t$ is warped using the predicted scene flow $\mathcal{SF}_t$ to obtain the estimated point cloud $\mathcal{PL}_{t+1}^{\omega}$ at time $t+1$.
The $\mathcal{L}_C$ loss function is designed to calculate the chamfer distance between $\mathcal{PL}_{t+1}^{\omega}$ and $\mathcal{PL}_{t+1}$. The formula is described as:
\begin{equation}
\begin{aligned}
    \mathcal{L}_C (\mathcal{PL}_{t+1}^{\omega}, \mathcal{PL}_{t+1})= \sum\limits_{\mathit{pl}_{2}^{\omega} \in \mathcal{PL}_{t+1}^{\omega} } \mathop{ min} \limits_{\mathclap{\mathit{pl}_{2} \in \mathcal{PL}_{t+1}}}{\left\| {\mathit{pl}_{2}^{\omega} - \mathit{pl}_{2}} \right\|_{2}^{2}} + \\
 \sum\limits_{\mathit{pl}_{2} \in \mathcal{PL}_{t+1}}\mathop {min}\limits_{\mathclap{\mathit{pl}_{2}^{\omega} \in \mathcal{PL}_{t+1}^{\omega}}}{\left\| {\mathit{pl}_{2}^{\omega} - \mathit{pl}_{2}} \right\|_{2}^{2}},
\end{aligned}
\label{eq:chamfer}
\end{equation}
where $\left\| \cdot \right\|_{2}^{2}$ represents the operation of Euclidean distance. The design of smoothness constraint $\mathcal{L}_{SC}$ is inspired by the a priori knowledge of smooth scene flow in real-world local space,
\begin{equation}
\mathcal{L}_{SC} = \sum\limits_{sf_i \in \mathcal{SF}} \frac{1}{|R(sf_i)|} \sum\limits_{sf_j \in R(sf_i)} \left\| sf_i - sf_j \right\|_{2}^{2},
\label{eq:sc}
\end{equation}
where $R(sf_i)$ means the set of all scene flow in the local space around $pl_i$. $|R(sf_i)|$ represents the number of points in $R(sf_i)$. Similar to $\mathcal{L}_{C}$, the goal of Laplacian regularization $\mathcal{L}_{LR}$ is to make the Laplace coordinate vectors of the same position in $\mathcal{PL}_{t+1}^{\omega}$ and $\mathcal{PL}_{t+1}$ consistent. The Laplace coordinate vector $o(\mathit{pl}_{2}^{\omega})$ of the point in $\mathcal{PL}_{t+1}^{\omega}$ is calculated as follows:
\begin{equation}
    o(\mathit{pl}_{2}^{\omega})= \frac{1}{|R(\mathit{pl}_{2}^{\omega})|}\sum\limits_{\mathit{pl}_{2}^{\omega} \in R(\mathit{pl}_{2}^{\omega}) } (\mathit{pl}_{2}^{\omega} - \mathit{pl}_{2}),
\end{equation}
where $R(\mathit{pl}_{2}^{\omega})$ represents the set of points in the local space around $\mathit{pl}_{2}^{\omega}$ and $|R(\mathit{pl}_{2}^{\omega})|$ is the number of points in $R(\mathit{pl}_{2}^{\omega})$. $\overline{o}$ is the interpolated Laplace coordinate vector from $\mathcal{PL}_{t+1}$ at the same position as $\mathit{pl}_{2}^{\omega}$ by using the inverse distance weight. $\mathcal{L}_{LR}$ is described as:
\begin{equation}
    \mathcal{L}_{LR}=\sum\limits_{\mathit{pl}_{2}^{\omega} \in \mathcal{PL}_{t+1}^{\omega}}\|o(\mathit{pl}_{2}^{\omega})-\overline{o}(\mathit{pl}_{2})\|.
\label{eq:laplace}
\end{equation}

Inspired by the coupling relationship between depth and pose in unsupervised depth pose estimation tasks \cite{bian2019unsupervised}, we propose a disparity consistency loss $\mathcal{L}_{DC}$. Specifically, each point on the first frame image is warped into the second frame by an estimated 3D scene flow, and the disparity or depth values from the warped points and the points in the real second frame should be the same. The disparity consistency loss is specifically described as:
\begin{equation}
    \mathcal{L}_{DC}= \mu (\mathit{Bl} \left[ \mathit{Pj}(\mathcal{PL}_{t+1}^{\omega}), D_{t+1} \right] - \mathit{Pj}(\mathcal{PL}_t)).
\label{eq:DepthC}
\end{equation}
where  $D_{t+1}$ represents the depth map at frame $t+1$. $Pj (\cdot)$ means the projection of the point cloud onto the image using the camera internal parameters. $Bl [\cdot]$ means the index of bilinear interpolation. $\mu(\cdot)$ means averaging over the tensor.

The overall loss of the scene flow estimator is as follow:
\begin{equation}
    \mathcal{L}=\sum\limits_{l=l_0}^L \Lambda_l (\lambda_1\mathcal{L}_C^l+\lambda_2\mathcal{L}_{SC}^l+\lambda_3\mathcal{L}_{LR}^l +\lambda_4\mathcal{L}_{DC}^l).
    \label{eq:Total_loss}
\end{equation}

The loss of the $l$-th level is a weighted sum of four losses. The total loss $\mathcal{L}$ is a weighted sum of the losses at each level. $\Lambda_l$ represents the weight of the loss in the $l$-th level.

\setlength{\tabcolsep}{1.4mm}
\begin{table*}[t]
    %\vspace{-5mm}
	\begin{center}
		\caption{All methods are evaluated on sfKITTI \cite{sfkitti}. ``Full'' and ``Self'' represent the supervised training and the self-supervised training, respectively. ``Stereo'', ``Mono'', and ``Points'' represent stereo images, monocular images, and point clouds, respectively. ``$\downarrow$'' and ``$\uparrow$'' are used to help the reader understand whether a larger or smaller value of the metric is better. The network is trained on the LiDAR point cloud/pseudo-LiDAR point cloud denoted as ``L'' and ``PL'', respectively. ``ft'' represents fine-tuning the model on sfKITTI, where the model is trained on 100 frames in sfKITTI and is evaluated on the remaining 42 frames.}
		\label{table:kitti}
		\begin{tabular}{c|c|c|c|cccc|cc}
			\toprule
			Methods & Training Set &  Sup.  &Input  & EPE3D($m$)$\downarrow$ & Acc3DS$\uparrow$  & Acc3DR$\uparrow$ &  Outliers3D$\downarrow$ & EPE2D($px$)$\downarrow$ & Acc2D$\uparrow$ \\ \midrule
			 FlowNet3 \cite{ref43} & FlyingC, FT3D  &  Full  & Stereo   & 0.9111                      & 0.2039             & 0.3587     &0.7463  & 5.1023 & \bf0.7803 \\ 
			FlowNet3D \cite{ref4} & FT3D   &  Full  & Points & 0.1767 & 0.3738 & 0.6677 & 0.5271 & 7.2141 & 0.5093 \\ \midrule
			
% 			 ICP \cite{ref42} & No &  No  & Points &   0.5181 & 0.0669 & 0.1667 & 0.8712 & 27.6752 & 0.1056  \\
% 			Ego-motion \cite{ref44} & FT3D &  Self & Points   & 0.4154 & 0.2209 & 0.3721 & 0.8096 & 15.0605 & 0.3162 \\
			Pontes et al. \cite{ref41} & FT3D &   Self  & Points & 0.1690 & 0.2171 & 0.4775 & --- & --- & ---  \\
			PointPWC-Net \cite{ref16} & FT3D  &  Self  & Points  & 0.2549 & 0.2379 & 0.4957 & 0.6863 & 8.9439 & 0.3299 \\
% 			PointPWC-Net \cite{ref16} & odKITTI  & Self & Points   &  0.6067 & 0.0202   & 0.0900   & 0.9286  & 25.0073   & 0.0756  \\
		    PointPWC-Net \cite{ref16} & FT3D, odKITTI & Self & Points   & 0.1699 & 0.2593   & 0.5718   & 0.5584       & 7.2800    & 0.3971 \\
		    Self-Point-Flow \cite{ref20} & FT3D & Self  & Points  & 0.1120 & \bf0.5276 & 0.7936 & 0.4086 & --- & ---  \\
% 			Mittal et al. \cite{ref27} & FT3D & Self  & Points   & 0.1220 & 0.2537 
% 			& 0.5785 & --- & --- & --- \\
			Mittal et al. (ft) \cite{ref27} & FT3D, sfKITTI & Self  & Points   & 0.1260 & 0.3200
			& 0.7364 & --- & --- & --- \\
            SFGAN (ft) \cite{wang2022sfgan} & FT3D, sfKITTI & Self  & Points & 0.1020 & 0.3205 & 0.6854 & 0.5532
            & --- & --- \\
% 			FLOT \cite{ref19} & FT3D  & Full & Points   &   0.0560 & 0.7550	& 0.9080	& 0.2420	& --- & ---  \\ 
			Ours PL (with Pre-train) & FT3D, odKITTI  & Self & Stereo   &  0.1103 & 0.4568 & 0.7412 & 0.4211  & 4.9141 & 0.5532 \\ 
			Ours PL (with Pre-train) & FT3D, odKITTI  & Self & Mono   &  \bf 0.0955 & 0.5118	& \bf0.7970 	& \bf0.3812	& \bf4.2671	& 0.6046  \\ 
			\midrule \midrule
			
			Ours L (w/o Pre-train) & odKITTI (64-beam)  & Self & Points   &  0.6067 & 0.0202   & 0.0900   & 0.9286       & 25.0073   & 0.0756  \\
			Ours L (w/o Pre-train) & DurLAR (128-beam)  & Self & Points   &  0.5078 & 0.0185   & 0.1026   & 0.9591       & 21.1068   & 0.1034  \\
			Ours PL (w/o Pre-train) & odKITTI & Self & Stereo   &  0.2179 &   0.2721  &  0.4616 &  0.6572  & 8.0812 &  0.3361  \\
			
			Ours L (with Pre-train) & FT3D, odKITTI (64-beam) & Self & Points   & 0.1699 & 0.2593   & 0.5718   & 0.5584       & 7.2800    & 0.3971 \\
			Ours L (with Pre-train) & FT3D, DurLAR (128-beam) & Self & Points   & 0.1595	& 0.2494	& 0.6318	& 0.5578	& 7.1517	& 0.3986  \\
			Ours PL (with Pre-train) & FT3D, odKITTI &  Self & Stereo   & 0.1103 & 0.4568 & 0.7412 & 0.4211  & 4.9141 & 0.5532 \\
			Ours PL (with Pre-train) & FT3D, odKITTI & Self & Mono   &  \bf 0.0955 & \bf0.5118	& \bf0.7970 	& \bf0.3812	& \bf4.2671	& 0.6046 \\
			\bottomrule
		\end{tabular}
	\end{center}
\end{table*}

\setlength{\tabcolsep}{0.5mm}
\begin{table*}[t]
    \vspace{-5mm}
	\begin{center}
		\caption{Quantitative results of scene flow estimation models evaluated on different real-world datasets. All networks learn the scene flow in a self-supervised approach and are not fine-tuned on lidarKITTI, Argoverse or nuScenes.}
		\label{table:dataset}
		%\resizebox{\textwidth}{8mm}{4.858985e-01;Outlier;8.547710e-01;ACC3DR;2.669922e-01;ACC3DS;8.219821e-02
		\begin{tabular}{c|cccc|cccc|cccc}
			\toprule
			Dataset & \multicolumn{4}{c}{lidarKITTI \cite{sfkitti}} & \multicolumn{4}{c}{Argoverse \cite{argoverse}} & \multicolumn{4}{c}{nuScenes \cite{nuscenes}} \\ \midrule
			Metrics  & EPE3D$\downarrow$ & Acc3DS$\uparrow$  & Acc3DR$\uparrow$ &  Outliers3D$\downarrow$ & EPE3D$\downarrow$ & Acc3DS$\uparrow$  & Acc3DR$\uparrow$ &  Outliers3D$\downarrow$ & EPE3D$\downarrow$ & Acc3DS$\uparrow$  & Acc3DR$\uparrow$ &  Outliers3D$\downarrow$ \\  \midrule
			PointPWC-Net \cite{ref16} & 1.1944 & 0.0384	& 0.1410 & 0.9336 & 0.4288 & 0.0462 & 0.2164 & 0.9199 & 0.7883 & 0.0287 & 0.1333 & 0.9410 \\ 
			Mittal et al. (ft) \cite{ref27} & 0.9773 & 0.0096 & 0.0524 & 0.9936 & 0.6520 & 0.0319 & 0.1159 & 0.9621 & 0.8422 & 0.0289 &  0.1041 & 0.9615 \\ 
			FLOT (Full) \cite{ref19} & 0.6532 & 0.1554 &0.3130 &0.8371& \bf0.2491& 0.0946 & 0.3126 & 0.8657 &
			0.4858& 0.0821& 0.2669 & 0.8547 \\ 
% 			OGSF (Full) \cite{ref19} & 0.4414 & 0.2821 & 0.4792 & 0.7130
% 			& 0.3248  & 0.4540  & 0.2567 & 0.7809
% 			&0.5368 & 0.1402 & 0.3081 & 0.8569 \\
            DCA-SRSFE \cite{jin2022deformation} &  0.5900 &0.1505 &0.3331   &0.8485
			&  0.7957 &0.0712 &0.1468  & 0.9799 
			&0.7042 & 0.0538 & 0.1183 & 0.9766 \\
			Ours (Stereo) & 0.5265 & 0.1732 & 0.3858 & 0.7638 & 0.2690 & 0.0768 & 0.2760 & 0.8440 
			& 0.4893 & 0.0554 & 0.2171 & 0.8649 \\ 
			Ours (Mono) & \bf0.4908 & \bf0.2052 & \bf0.4238 & \bf 0.7286 & 0.2517 & \bf0.1236 & \bf0.3666 & \bf0.8114 
			& \bf0.4709 & \bf0.1034 & \bf0.3175 & \bf0.8191 \\ 

			\bottomrule
		\end{tabular}
	\end{center}
\end{table*}

\section{Experiments}\label{Experiments}
\subsection{Experimental Details}
\subsubsection{Training Settings}
The proposed algorithm is written in Python and PyTorch and runs on Linux. On a single NVIDIA TITAN RTX GPU, we train for 40 epochs. The initial learning rate is set to 0.0001 and the learning rate decreases by 50\% every five training epochs. The batch size is set to 4. The generated pseudo-LiDAR is randomly sampled to 4096 points as input of the scene flow estimator. With the same parameter settings as PointPWC-Net, there are four levels of the feature pyramid in the scene flow estimator in this paper. In Eq. (\ref{eq:Total_loss}), the first level weight $\Lambda_{l_0}$ to the fourth level weight $\Lambda_{l_3}$ are 0.02, 0.04, 0.08, and 0.16. The self-supervised loss weights are $\lambda_1 = 1.0$, $\lambda_2 = 0.2$, $\lambda_3 = 0.2$, and $\lambda_4 = 1.0$, respectively.

Depth annotations in the synthesized dataset  \cite{ref24} are used to supervise the depth estimator, similar to Pseudo-lidar++ \cite{ref30}. The pre-trained depth estimator is fine-tuned utilizing LiDAR points from the KITTI \cite{ref40} as sparse ground truth, as indicated in Eq. \ref{eq:Z_Z}. During the scene flow estimator training stage, the depth estimator weights will be fixed. The scene flow estimator is first pre-trained on FT3D \cite{ref24} with self-supervision method. Stereo images from the 00-09 sequence of the KITTI odometry (odKITTI) \cite{ref40} are selected to train our scene flow estimation model. To further improve the applicability of the method, we also explored a framework for monocular vision estimation of 3D scene flow, where the depth estimator uses the advanced monocular depth model AdaBins \cite{bhat2021adabins}.

To further demonstrate the denseness advantage of the pseudo-LiDAR point cloud proposed in section \ref{l_pl}, the scene flow estimator is trained on a denser LiDAR point clouds from the high-fidelity 128-Channel LiDAR Dataset (DurLAR) \cite{durlar}. To be fair, we perform the same processing as PointPWC-Net \cite{ref16} for the LiDAR point cloud in DurLAR. The results are presented at the bottom of Table \ref{table:kitti}.

\subsubsection{Evaluation Settings}
Following PointPWC-Net \cite{ref16}, we evaluate the model performance on the KITTI Scene Flow dataset (sfKITTI) \cite{sfkitti}, where sfKITTI is obtained through 142 pairs annotations of disparity maps and optical flow. The lidarKITTI \cite{sfkitti}, with the same 142 pairs as sfKITTI, is generated by projecting the LiDAR point clouds of 64-beam onto the images. The 142 frame scenes are all used as test samples.  Following Pontes et al. \cite{ref41}, we also evaluate the generalizability of the proposed method on two real-world datasets, Argoverse \cite{argoverse} (containing 212 test samples) and nuScenes \cite{nuscenes} (containing 310 test samples). Different from Pontes et al. \cite{ref41}, our methods have not accessed any data from Argoverse \cite{argoverse} and nuScenes \cite{nuscenes} in the training process. All methods in the table evaluate the performance of the scene flow directly on Argoverse and nuScenes. To be fair, we use the same evaluation metrics as PointPWC-Net \cite{ref16}.

%\subsection{Evaluation Metrics}
%We focus on estimating 3D scene flow and report the evaluation results on KITTI Scene Flow 2015 dataset \cite{ref40} without fine-tuning. We adopt the same evaluation metrics as \cite{ref16}, which contain four 3D metrics (EPE3D ($m$), Acc3DS, Acc3DR, and Outliers3D) and two 2D metrics (EPE2D ($px$) and Acc2D).

\subsection{Results}
\begin{figure*}[t]
	\centering
	\includegraphics[scale=0.54]{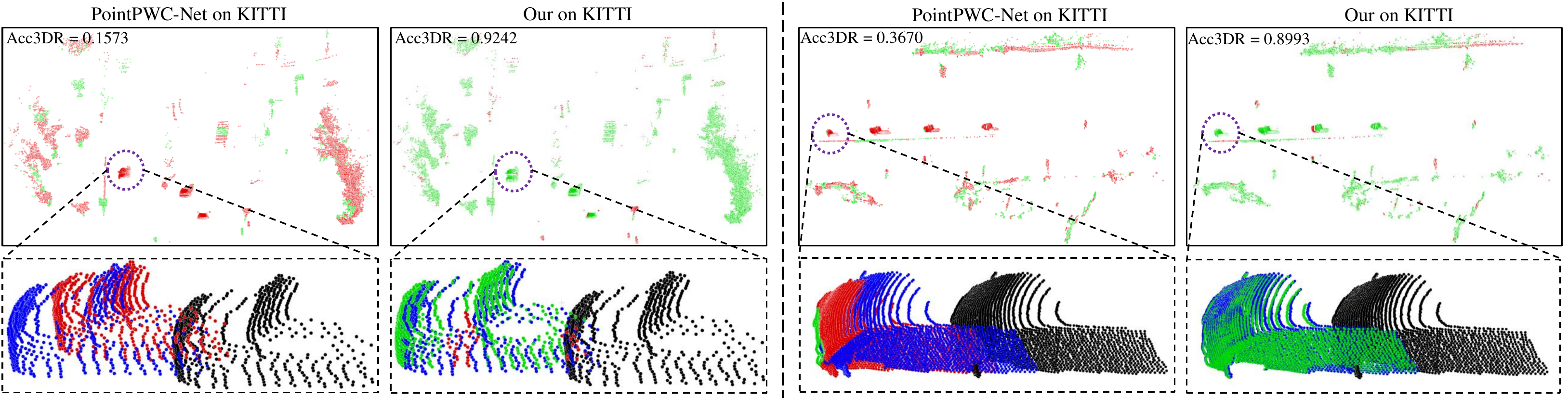}
	%\includegraphics[scale=0.545]{Pseudo12.png}
	%\vspace{-0.5mm}
	\caption{\textbf{Visualization of the results of scene flow estimation on sfKITTI \cite{sfkitti}.} PointPWC-Net \cite{ref16} is used as the baseline. The predicted point cloud $\mathcal{PC}_{t+1}^{\omega}$ is synthesized by warping the point cloud $\mathcal{PC}_t$ of the first frame (present as black points) to the point cloud of the second frame using the predicted scene flow $\mathcal{SF}$, that is, $\mathcal{PC}_t+\mathcal{SF}$. Prediction point with error $<$ 0.1$m$ or relative error $<$ 10\% are classified as correct point. (measured by Acc3DR). The estimated accurate points and inaccurate points are shown in green and red, respectively. The real point cloud of the second frame is represented as blue. To show the effect of comparing the predicted point cloud with the ground truth point cloud, some detail in the 3D dynamic scene is selected and zoomed in.}
	\label{fig:errorAcc}
\end{figure*}
% \begin{figure*}[t]
% 	\centering
% 	\includegraphics[scale=0.088]{Pseudo3.png}
% 	\caption{\textbf{Visualization of the accuracy of scene flow estimation on sfKITTI.} Points with correct prediction are shown as green and points with incorrect prediction are shown as pink in the predicted point cloud of the second frame. Prediction point with error $>$ 0.3$m$ or relative error $>$ 10\% are classified as error point. (measured by Outliers3D).}
% 	\label{fig:errorOut}
% \end{figure*}

%, and non-learning method ICP \cite{ref42} In addition, although our method mainly emphasizes 3D scene flow estimation, we still outperform FlowNet3 on the 2D metric EPE2D. Our method outperforms even the supervised method FlowNet3D \cite{ref4} on all metrics. 

Table \ref{table:kitti} and table \ref{table:dataset} gives the quantitative results of our method evaluated at sfKITTI \cite{sfkitti}. The accuracy of our method is substantially ahead of supervised learning methods FlowNet3 \cite{ref43}, FlowNet3D \cite{ref4}, and FLOT \cite{ref19}. Compared with the self-supervised learning method \cite{ref16,ref41,ref27,wang2022sfgan,ref20,jin2022deformation} based on point clouds, learning 3D scene flow on pseudo-LiDAR from real-world images demonstrates an impressive effect. Compared to PointPWC-Net \cite{ref16}, our method improves over 45\% on EPE3D, Acc3DS and EPE2D, and improves over 30\% on Acc3DR, Outliers and Acc2D, which is a powerful demonstration that Chamfer loss and Laplacian regularization loss can be more effective on pseudo-LiDAR. Our method without fine-tuning still outperforms the results of Mittal et al. (ft) \cite{ref27} fine-tuning on sfKITTI.

From the bottom of Table \ref{table:kitti}, the 3D scene flow estimator is trained from scratch on a synthesized dataset \cite{ref24}, LiDAR point clouds \cite{ref40,durlar}, and real-world stereo/monocular images \cite{ref40}, respectively. It is obvious that self-supervised learning works best on pseudo-LiDAR point clouds estimated from images. This confirms that our method avoids the limitation of estimating the scene flow on the synthesized data set to a certain extent. It also confirms that our method avoids the weakness of learning point motion from LiDAR data. Furthermore, in the last row of Table \ref{table:kitti}, the model trained on FT3D \cite{ref24} serves as a prior guide for our method. As we have been emphasizing, pseudo-LiDAR point clouds will stimulate the potential of self-supervised loss to a greater extent.

\begin{figure}[t]
	\centering
	\includegraphics[scale=0.39]{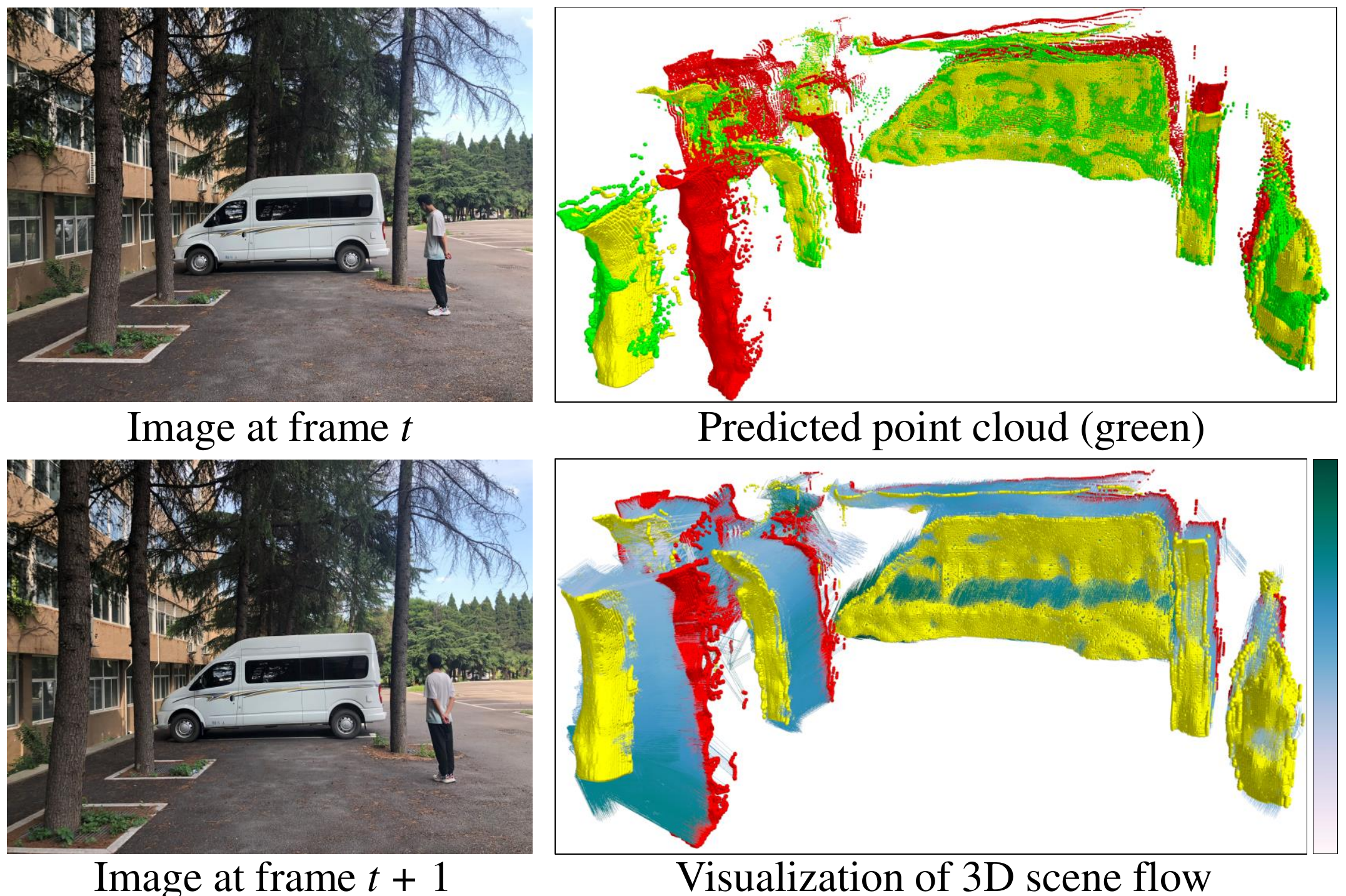}
	%\vspace{-1mm}
	\caption{\textbf{Visualization of the results of our designed network for estimating 3D scene flow in realistic scenes.}}
	\label{fig:realworld}
\end{figure}

The accuracy of learning 3D scene flow on the 128-beam LiDAR signals \cite{durlar} is improved compared to the 64-beam LiDAR signals. According to the sentence according to the quantitative results of Table \ref{table:kitti} and Table \ref{table:dataset}, the proposed framework for learning 3D scene flow from pseudo-LiDAR signals still presents greater advantages. To qualitative demonstrate the effectiveness of our method, some visualizations are shown in Fig. \ref{fig:errorAcc}. Compared to PointPWC-Net \cite{ref16}, the estimated points by our method are mostly correct points on the Acc3DR metric. From the details in Fig. \ref{fig:errorAcc}, the point clouds estimated from our method overlap well with the ground truth point clouds, confirming the reliability of our method. Finally, the methods in this paper also show excellent perceptual performance in the real world as shown in Figure \ref{fig:realworld}.

%$w/o$ outliers
\setlength{\tabcolsep}{1mm}
\begin{table}[t]
	\begin{center}
		\caption{``Edges'' represents the range that specifies pseudo-LiDAR point cloud to remove more estimated error points. ``Outliers'' represents the elimination of outlier points within pseudo-LiDAR point clouds.}
		\label{table:ab2}
		\begin{tabular}{ccc|ccc}
			\toprule
			 edges   & outliers  & $\mathcal{L}_{DC}$  & EPE3D(m)$\downarrow$ & Acc3DR$\uparrow$ & EPE2D(px)$\downarrow$ \\ \midrule
			 $\times$ & $\times$ & $\times$
			& 0.2655  & 0.3319  & 11.4530\\ 
			\checkmark  & $\times$ & $\times$
			&0.1191 & 0.7181  & 5.3741 \\
			\checkmark     & \checkmark & $\times$
			&   0.1156 & 0.7298 & 5.0802 \\ 
			\checkmark     & \checkmark & \checkmark 
			&   \bf 0.1103 & \bf0.7412 & \bf4.9141 \\ 
			\bottomrule
		\end{tabular}
	\end{center}
\end{table}
\setlength{\tabcolsep}{2mm}
\begin{table}[t]
	\begin{center}
		\caption{Ablation studies on outlier removal. The effects of two parameters, the number $m$ of selected nearest points and the scaling factor $\alpha$, on 3D scene flow learning are explored.}
		\label{table:ab3}
		\begin{tabular}{c|c|cccc|cc}
			\toprule
			 $m$   & $\alpha$    & EPE3D($m$)$\downarrow$ & Acc3DS$\uparrow$ & EPE2D(px)$\downarrow$ \\ \midrule
			4   &  2 
			& 0.1155 & 0.4224   & 5.3325\\
			8   &  1 
			& 0.1188 & 0.4044   & 5.3622\\ 
			8   &  2 
			&   \bf 0.1103 & \bf0.4568 & 4.9141\\ 
			8   &  4 
			& 0.1111 & 0.4422   & \bf4.8438\\ 
			16         & 2
			& 0.1261 & 0.4237   & 5.5634 \\
			 \bottomrule
		\end{tabular}
	\end{center}
\end{table}

% In addition, as shown in Fig. \ref{fig:errorOut}, estimated points with large errors are less by our method. The proposed method spends most of its time on depth estimation and pseudo-LiDAR signal processing during the training process. Therefore, w

We test the runtime on a single NVIDIA TITAN RTX GPU. PointPWC-Net takes about 0.1150 seconds on average to perform a training step while our method takes 0.6017 seconds. We consider saving the pseudo-LiDAR point cloud from the depth estimation and enabling the scene flow estimator to learn the scene flow from the saved point cloud. After saving the pseudo-LiDAR signals, the training time is greatly reduced and achieves the same time consumption (about 0.1150 seconds) as PointPWC-Net.

%We only evaluate the scene flow estimator in our evaluation experiments, where the scene flow estimator estimates the 3D scene flow from a pair of 3D point clouds. It is worth noting that, the proposed model takes 0.1040 seconds to perform an inference step, as with PointPWC-Net. Pseudo-LiDAR point cloud generated in the depth map contains many bad artificial points around the objects, which cannot provide useful information for scene flow estimation. The elimination of the redundant points is crucial for scene flow estimation. Some ablation studies on the processing of pseudo-LiDAR are performed. 

\subsection{Ablation Studies}
The edge points of the whole scene and the outliers within the scene are the two factors that we focus on. In Table \ref{table:ab2}, the experiments demonstrate that the choice of a suitable scene range and the elimination of outlier points both facilitate the learning of 3D scene flow. In the ablation study, the proposed disparity consistency loss $\mathcal{L}_{DC}$  is verified to be very effective for learning 3D scene flow. A point in the point cloud whose distance from its nearest point exceeds a distance threshold $d_{max}$ is considered as an outlier, where $d_{max}$ is calculated by Eq. (\ref{eq:Out}). The probability that a point in the point cloud is considered as an outlier is determined by the number of selected points $m$ and the standard deviation multiplier threshold $\alpha$. The experiments in Table \ref{table:ab3} show the best results for elimination of outlier points when $m$ is 8 and $\alpha$ is 2. 

%The method of directly recovering 3D scene flow from stereo images of consecutive frames is proposed. Besides, there are many inaccurate points in the pseudo-LiDAR point cloud due to the error in depth estimation, and these inaccurate points are filtered out as much as possible to benefit the scene flow estimation. In addition, LiDAR is a very expensive component in the field of robotics and autonomous driving. Replacing some of the LiDAR tasks with cameras is meaningful work for reducing hardware costs. which contributes to the advancement of estimating 3D scene flow using pseudo-LiDAR. 
\section{Conclusion}\label{Conclusion}
The method in this paper achieves accurate perception of 3D dynamic scenes on 2D images. The pseudo-LiDAR point cloud is used as a bridge to compensate for the disadvantages of estimating 3D scene flow from LiDAR point clouds. The points in the pseudo-LiDAR point cloud that affect the scene flow estimation are filtered out. In addition, disparity consistency loss was proposed and achieved better self-supervised learning results. The evaluation results demonstrate the advanced performance of our method in the real world datasets.

\ifCLASSOPTIONcaptionsoff
  \newpage
\fi

\bibliographystyle{IEEEtran}
\bibliography{IEEEabrv}

\end{document}